\documentclass[conference]{IEEEtran}
\IEEEoverridecommandlockouts
\usepackage{cite}
\usepackage{amsmath,amssymb,amsfonts}
\usepackage{algorithmic}
\usepackage{graphicx}
\usepackage{textcomp}
\usepackage[table]{xcolor}

\usepackage{multirow}

\usepackage{amsthm}
\usepackage{algorithm}
\usepackage{pifont}
\usepackage{colortbl}
\usepackage{array}
\usepackage{booktabs}

\usepackage{tabularx}
\usepackage[caption=false]{subfig}
\usepackage{hyperref}

\def\BibTeX{{\rm B\kern-.05em{\sc i\kern-.025em b}\kern-.08em
    T\kern-.1667em\lower.7ex\hbox{E}\kern-.125emX}}
    
\begin{document}

\title{HyperDiff: Hypergraph Guided Diffusion Model for 3D Human Pose Estimation}


\author{
    \IEEEauthorblockN{ Bing Han, Yuhua Huang$^{\ast}$\thanks{*Corresponding author}, and Pan Gao$^{\ast}$}
    \IEEEauthorblockA{
        \textit{Nanjing University of Aeronautics and Astronautics, Nanjing, China} \\
        \{icehan@nuaa.edu.cn, hyuhua2k@163.com, gaopan.1005@gmail.com\} 
    }
}

\maketitle

\begin{abstract}
Monocular 3D human pose estimation (HPE) often encounters challenges such as depth ambiguity and occlusion during the 2D-to-3D lifting process. Additionally, traditional methods may overlook multi-scale skeleton features when utilizing skeleton structure information, which can negatively impact the accuracy of pose estimation. To address these challenges, this paper introduces a novel 3D pose estimation method, HyperDiff, which integrates diffusion models with HyperGCN. The diffusion model effectively captures data uncertainty, alleviating depth ambiguity and occlusion. Meanwhile, HyperGCN, serving as a denoiser, employs multi-granularity structures to accurately model high-order correlations between joints. This improves the model's denoising capability especially for complex poses. Experimental results demonstrate that HyperDiff achieves state-of-the-art performance on the Human3.6M and MPI-INF-3DHP datasets and can flexibly adapt to varying computational resources to balance performance and efficiency. Code is released at https://github.com/IHENL/HyperDiff
\end{abstract}


\section{Introduction}
Monocular 3D Human Pose Estimation (HPE) aims to predict the positions of human joints in 3D space from 2D images or video sequences. As a fundamental task for various downstream visual applications, it plays a key role in the fields such as virtual reality~\cite{chen2019so,chen2021model,chen2021joint}, human-computer interaction~\cite{wang2021synthesizing, hassan2021populating,ng2020you2me}, and autonomous driving~\cite{lu2022kemp}. Traditional methods typically decompose the 3D HPE process into two stages: (1) 2D joint estimation, where existing 2D keypoint detectors identify 2D keypoints~\cite{newell2016stacked,wei2016convolutional,chen2018cascaded,sun2019deep} from RGB images, and (2) 2D-to-3D lifting, where the detected 2D keypoints are mapped into 3D poses. This paper  focuses  primarily on the second stage—2D-to-3D lifting—aiming to estimate accurate 3D poses from 2D keypoints.

Accurately estimating 3D joint positions from 2D coordinates remains challenge due to inherent depth ambiguities and frequent self-occlusions. Additionally, the high variability in human motion and body configurations in monocular data introduces substantial uncertainty in the lifting process. To address these issues, an increasing number of studies have explored the use of diffusion models~\cite{ho2020denoising, song2020denoising}. Diffusion models~\cite{shan2023diffusion,xu2024finepose,cai2024disentangled} gradually add noise to ground truth data and remove it during generation, effectively modeling data uncertainty and alleviating depth ambiguities and occlusion problems.
Nevertheless, existing diffusion-based methods usually flatten the 2D pose before inputting it into the model, which limits the full utilization of structural skeleton information. Moreover, while human skeleton can be naturally represented using graph structures~\cite{li2025graphmlp,shi2019two,yan2018spatial}, and Graph Convolutional Networks (GCNs)~\cite{bruna2013spectral,kipf2016semi} suitably capture joint interactions, most existing GCN-based methods~\cite{hu2021conditional,yu2023gla,li2025graphmlp,zhao2022graformer} focus solely on joint-level information, neglecting the broader physiological structure.

This study presents a new diffusion model to model data uncertainty, where it synergistically uses HyperGCN~\cite{Fey/Lenssen/2019,yadati2019hypergcn, wei2021dynamichypergraphconvolutionalnetworks,zhu2022selective} to enhance the capability of the model to capture and preserve detailed information about skeleton structure. Specifically, we train a spatial HyperGCN to denoise contaminated 3D poses conditioned on 2D keypoints. Combined with a reconstruction objective, the denoiser implicitly captures spatial correlations between the joints within the frame.
Moreover, we enhance the denoiser by introducing finer-grained spatial scales, segmenting the human skeleton into three granularities~\cite{feng2019hypergraph,zhu2022selective}: \emph{joint-scale graph}, \emph{part-scale hypergraph} and \emph{body-scale hypergraph}, as illustrated in Fig. 1. This segmentation enables more precise modeling of local joint structures, thereby improving the denoiser's performance.
\begin{figure}[t]
    \centering
        \includegraphics[width=\linewidth]{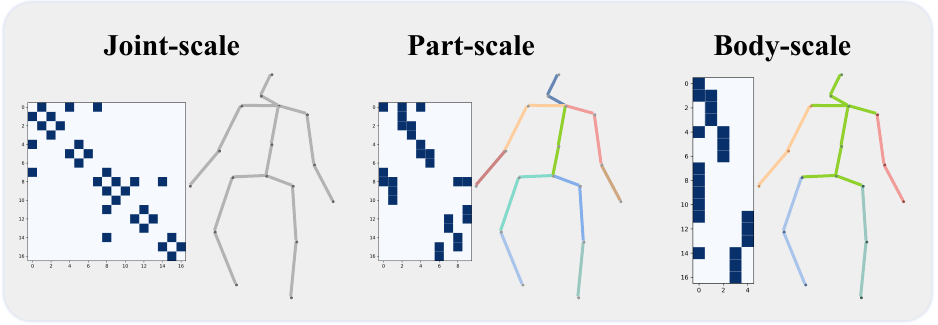} \\ 
    \vspace{-2mm}
	\caption{Graph and hypergraph based human skeleton represention.}
	\label{fig:motivation} 
    \vspace{-4mm}
\end{figure}
This novel diffusion-based HPE method is termed  \textbf{HyperDiff} (\textbf{Hyper}graph Guided \textbf{Diff}usion Model). By modeling high-order joint correlations without destroying the skeleton, HyperDiff yields more accurate 3D poses. In summary, the main contributions of this paper are:
\begin{itemize}
    \item We propose the HyperDiff framework, which leverages the HyperGCN as the denoiser for diffusion models. This effectively addresses uncertainties and depth ambiguities in 3D pose estimation.
    \item We further introduce a multi-granularity HyperGCN structure that optimizes high-order structural modeling and improves the denoiser's performance, thereby enhancing pose accuracy and structural expressiveness.
    \item HyperDiff achieves state-of-the-art results on the Human3.6M and MPI-INF-3DHP datasets. Additionally, it balances performance and efficiency under varying computational resource conditions by adjusting the number of denoising and iteration steps.
\end{itemize}

\section{Method}
\begin{figure*}[htp]
	\centering
		\includegraphics[width=\textwidth]{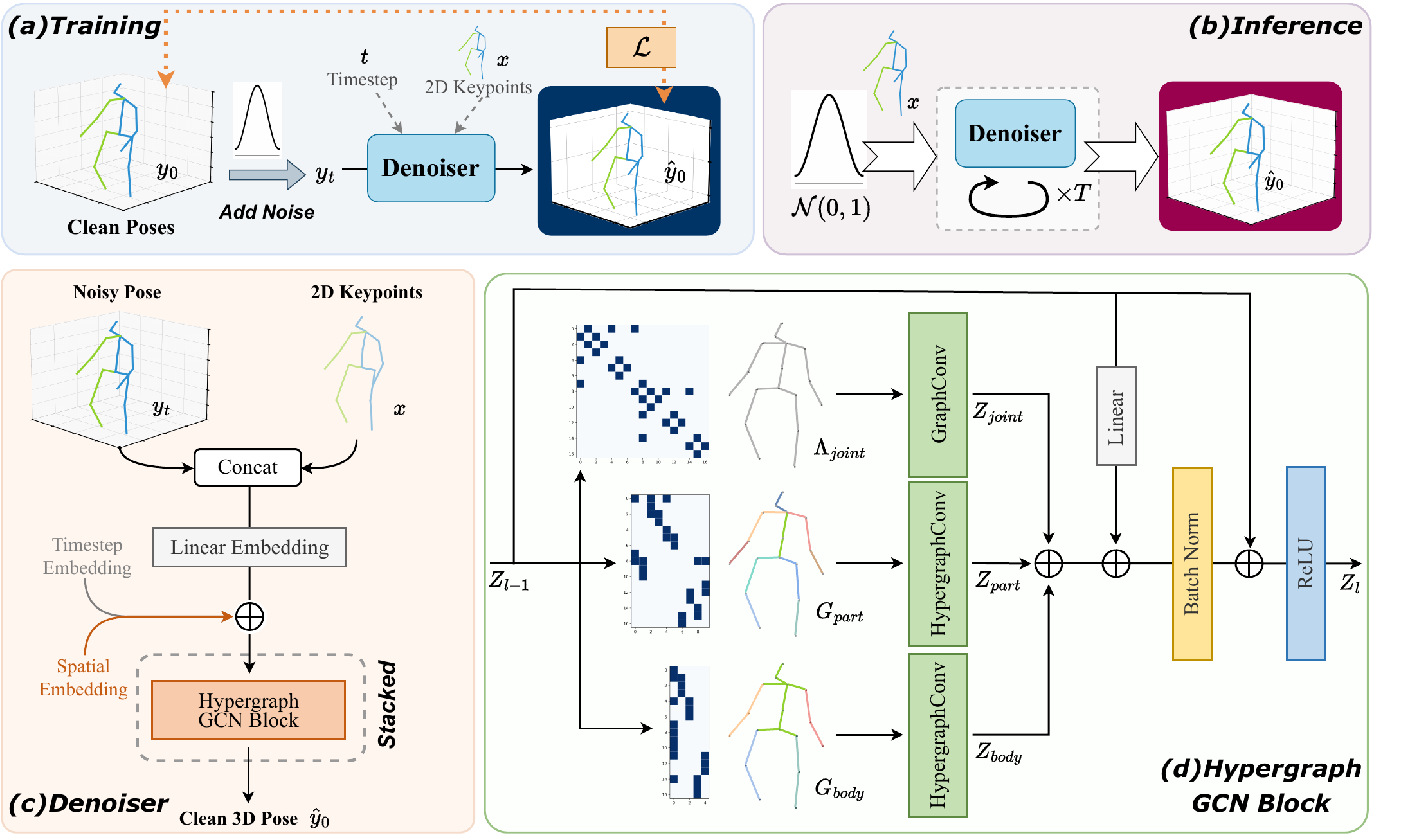} \\
        \vspace{-2mm}
	\caption{Overview of the proposed HyperDiff framework. (a) and (b) illustrate the training and inference processes. (c) shows the architecture of the denoiser, while (d) depicts the structure of the proposed Hypergraph GCN Block.}
	\label{fig:method_overview} 
    \vspace{-2mm}
\end{figure*}
\subsection{HyperGCN Strategy}
\label{sec:HyperGCN_strategy}
We first introduce the motivation for adopting the HyperGCN in this study. Traditional diffusion methods~\cite{shan2023diffusion} directly take 3D joint as input, lacking prior skeleton structure information. Due to the complex and dense relationships between joint pairs, modeling their dependencies is challenging, which complicates the optimization task. To address these issues, we consider that the human skeleton can be represented as a graph, and GCN can more effectively capture interactions between joints. However, most existing GCN-based methods rely on single-scale joint information to extract skeleton features, ignoring the valuable multi-scale contextual information. To this end, we construct two types of hypergraphs based on the dynamic chain structure of the human body: the part-scale hypergraph and the body-scale hypergraph, defined below:
\begin{equation}
   \begin{aligned}
        &p1=\{hip,spine,thorax\} \quad      p2=\{thorax,neck,head\} \\
        &p3=\{hip,rhip,rknee\}   \quad\quad p4=\{rknee,rfoot\} \\
        &p5=\{hip,lhip, lknee\}   \quad\quad p6=\{lknee,lfoot\} \\
        &p7=\{relbow,rwrist\} \quad\quad p8=\{lelbow,lwrist\} \\
        &p9=\{thorax,rshoulder,relbow\} \\
        &p10=\{thorax,lshoulder,lelbow\} \\
        &b1=\{hip,rhip,lhip,spine,thorax,neck,head,\\
        &\quad\quad lshoulder,rshoulder\}\\
        &b2=\{rhip,rknee,rfoot\} \quad b3=\{lhip,lknee,lfoot\}\\
        &b4=\{rshoulder,relbow,rwrist\}\\
        &b5=\{lshoulder,lelbow,lwrist\}\\
    \end{aligned}
\end{equation}

Here, $p_i$ represents the joints included in the hyperedge of the part-scale hypergraph $H_{part}$, and $b_i$ represents the joints included in the hyperedge of the body-scale hypergraph $H_{body}$. 

Specifically, the joint-scale graph focuses on the dependency  between joints, effectively modeling their direct connections and local dependencies, thereby capturing the local relationships between joints. The part-scale hypergraph further enhances the modeling of local joint dependencies by defining the relationships between different body parts. The body-scale hypergraph, on the other hand, considers broader global relationships and can capture the interdependencies between different body regions, thus providing a more comprehensive modeling of the human anatomy. This hierarchical division enables the model to accurately model dependencies between joints at both local and global scales. It also enhances the expressiveness of skeleton structures by capturing higher-order and more complex joint interactions. As a result, the precision and optimization of 3D pose estimation are improved.

\subsection{Diffusion-based Estimation Strategy}
\label{sec:method_diffusion_strategy}
In this section, we outline the overall diffusion-based strategy of HyperDiff. Diffusion models simulate data distributions via forward and backward processes. HyperDiff defines the forward process as gradually adding timestep-dependent Gaussian noise $ \epsilon \sim \mathcal{N}(0, I) $ to the ground truth 3D pose $ y_0 \in \mathbb{R}^{J \times 3} $. Following DDPM~\cite{ho2020denoising}, this process is expressed as:
\begin{equation}
y_t = \sqrt{\bar{\alpha}_t}y_0 + \epsilon\sqrt{1 - \bar{\alpha}_t} 
\end{equation}
where the timestep $t={1,..,T}$. $\bar{\alpha}_t := \prod_{s=0}^{t} \alpha_s$ and $\alpha_t := 1 - \beta_t$. $\{\beta_t\}_{t=1}^{T}$ is the cosine noise schedule.

In the reverse process, the noisy 3D pose $ y_t \in \mathbb{R}^{J \times 3} $ is input into a denoiser $ \mathfrak{D} $, which conditions on the 2D pose $ x \in \mathbb{R}^{J \times 2} $ and time step $ t $ to produce the denoised 3D pose $ \hat{y}_0 $:
\begin{equation}
\hat{y}_0 = \mathfrak{D}(y_t, x, t)
\label{eq:denoiser}
\end{equation}

During inference, we sample $ H $ initial noisy poses $ y_{0:H,t} $ from Gaussian noise $ \mathcal{N}(0, 1) $. These are denoised using the trained $ \mathfrak{D} $ to get $ \hat{y}_{0:H,0} $. Following DDIM~\cite{song2020denoising}, we generate the next iteration's noisy samples $ y_{0:H,t'} $ from $ \hat{y}_{0:H,0} $, which serve as inputs for the denoiser at time step $ t' $. This can be formalized as:

\begin{equation}
    \begin{aligned}
        y_{0:H,t'} &= \sqrt{\bar{\alpha}_{t'}}  \hat{y}_{0:H,0} + \epsilon_{t} \sqrt{1 + \bar{\alpha}_{t} - \sigma_{t}^2} + \sigma_{t} \epsilon
        \label{eq:denoised}
    \end{aligned}    
\end{equation}
where $t$ and $t'$ are the current and next timesteps, respectively, with $t$ ranging from $T$ to 1. $\epsilon \sim \mathcal{N}(0, I)$ is standard Gaussian noise independent of $y_{0:H,t}$ , and
\begin{equation}
    \begin{aligned}
        \epsilon_{t} &= (y_{0:H,t} - \sqrt{\bar{\alpha}_{t}}\hat{y}_{0:H,0}) / \sqrt{1-\bar{\alpha}_{t}}\\
        \sigma_{t} &= \sqrt{(1-\bar{\alpha}_{t'})/(1-\bar{\alpha}_{t}}) \cdot \sqrt{1-\bar{\alpha}_{t} / \bar{\alpha}_{t'}}
    \end{aligned}    
\end{equation}
where $\epsilon_{t}$ represent the noise at timestep $t$, and $\sigma_{t}$  are parameters that govern the degree of stochasticity in the diffusion process.

The iterative process is repeated $ K $ times, beginning at timestep $ T $. The timestep of each iteration is determined by $ t = T \cdot \left(1 - \frac{k}{K}\right) $, $k \in [0, K) $.

\subsection{Denoiser Architecture Design}
\label{sec:method_denoiser_architecture}

This section presents the denoiser architecture based on hypergraph convolution. As illustrated in Fig.~\ref{fig:method_overview}(c), given the input noisy pose $y_t \in \mathbb{R}^{J \times 3}$, we concatenate it with the corresponding 2D pose $x \in \mathbb{R}^{J \times 2}$, resulting in $y'_t \in \mathbb{R}^{J \times (3+2)}$. We then apply linear embedding to project the feature dimensions to $d_m$, adding spatial and time-step embeddings to obtain the embedded token $Z \in \mathbb{R}^{J \times d_m}$. Next, $Z$ is fed into the stacked hypergraph GCN block to learn multi-level spatial representations. Finally, a projection head is used to convert the features into the clean 3D pose $\hat{y}_0 \in \mathbb{R}^{J \times 3}$.

Recent studies have shown the effectiveness of GCN-based architectures for spatial representation modeling in inferring accurate 3D joint positions from 2D keypoints (\cite{zou2021modulated,zhao2022graformer,li2025graphmlp}). The standard graph convolution operation is defined as:
\begin{equation}
    GCN(Z)=\sigma(D^{-\frac{1}{2}}\widetilde{A}D^{-\frac{1}{2}}ZW)=\sigma(\Lambda ZW)
\end{equation}
where $\sigma$ is the activation function, $\widetilde{A} = A + I$ is the adjacency matrix with self-loops, $D$ is the degree matrix, and $W$ is the learnable weight matrix. To simplify the notation, we define $\Lambda = D^{-\frac{1}{2}} \widetilde{A} D^{-\frac{1}{2}}$ as the graph convolution kernel.

With the development of graph convolution, researchers have extended this operation to the hypergraph~\cite{bai2021hypergraph,yadati2019hypergcn}. Given the hypergraphs defined in \ref{sec:HyperGCN_strategy}, hypergraph convolution can aggregate skeleton information across multiple scales. The hypergraph convolution operation is defined as follows:
\begin{equation}
    HGCN(Z)=\sigma(D^{-\frac{1}{2}}_vHMD^{-1}_eH^TD^{-\frac{1}{2}}_vZW)=\sigma(GZW))
\end{equation}
where $H$ is the hypergraph's adjacency matrix, $D_e$ is the diagonal matrix of hyperedge degrees, $D_v$ is the diagonal matrix of vertex degrees, $W$ is the learnable node embedding weight matrix, and $M$ is the learnable hyperedge weight matrix, initially set as the identity matrix (i.e., all hyperedge weights are equal). For simplicity, we use $G = D_v^{-\frac{1}{2}} H M D_e^{-1} H^T D_v^{-\frac{1}{2}}$ as the hypergraph convolution kernel.

As shown in Fig.~\ref{fig:method_overview}(d), the hypergraph GCN block first passes the input feature $Z_{l-1} \in \mathbb{R}^{J \times d_m}$ through three convolution branches:
\begin{equation}
    \begin{aligned}
        Z_{joint}&= \sigma(\Lambda_{joint}Z_{l-1}W_{joint}) \\
        Z_{part}&=\sigma(G_{part}Z_{l-1}W_{part}) \\
        Z_{body}&=\sigma(G_{body}Z_{l-1}W_{body})
    \end{aligned}    
\end{equation}
where $Z_{joint}, Z_{part}$ and $Z_{body}$ represent the skeleton features obtained at different scales, $G_{part}$ and $G_{body}$ are the hypergraph convolution kernels at the part-scale and body-scale, and $W_{part}$ and $W_{body}$ are the learnable weight matrices. The features from these three branches are then fused using element-wise summation:
\begin{equation}
    Z=\alpha_{joint} \cdot Z_{joint}+\alpha_{part} \cdot Z_{part} + \alpha_{body} \cdot Z_{body}
\end{equation}
where $\alpha_{joint}, \alpha_{part}$ and $\alpha_{body}$ are learnable weights. This weighted fusion allows the model to comprehensively integrate information from local, part, and global scales, thereby capturing joint dependencies more effectively. Finally, $\text{ReLU}(Z_{l-1} + \text{BN}(Z + \text{Linear}(Z_{l-1})))$ produces $Z_l$, which contains enriched skeleton feature information, serving as the output of the hypergraph GCN block for subsequent operations.

\begin{table*}[t]
\caption{Comparison of the proposed method with state-of-the-art methods on Human3.6M using MPJPE. The top group uses detected 2D poses and the bottom group uses ground truth 2D poses as inputs. The best and second-best results are marked in \textcolor{red}{red} and \textcolor{blue}{blue}, respectively. $T, H, K$: the number of input frames, hypotheses, and iterations of D3DP~\cite{shan2023diffusion}. (\textdaggerdbl) denotes using visual cues.}
\setlength{\tabcolsep}{1.3mm} 
\centering
\scalebox{0.92}{
\begin{tabular}{l|cccccccccccccccc}
\hline
MPJPE       & Dir  & Disc & Eat  & Greet & Phone & Photo & Pose & Pur  & Sit  & SitD & Smoke & Wait & WalkD & Walk & WalkT & Avg  \\ \hline
MGCN\cite{zou2021modulated}          & 45.4 & 49.2 & 45.7 & 49.4 & 50.4 & 58.2 & 47.9 & 46   & 57.5 & 63   & 49.7 & 46.6 & 52.2 & 38.9 & 40.8 & 49.4 \\
Graformer\cite{zhao2022graformer} &45.2 & 50.8 & 48.0 & 50.0 & 54.9 & 65.0 & 48.2 & 47.1 & 60.2 & 70.0 & 51.6 & 48.7 & 54.1 & 39.7 & 43.1 & 51.8 \\
POT\cite{li2023pose} & 47.9 & 50   & 47.1 & 51.3 & 51.2 & 59.5 & 48.7 & 46.9 & 56   & 61.9 & 51.1 & 48.9 & 54.3 & 40   & 42.9 & 50.5 \\
DiffPose\cite{gong2023diffpose}     & \textcolor{blue}{42.8} & 49.1 & 45.2 & 48.7 & 52.1 & 63.5 & 46.3 & 45.2 & 58.6 & 66.3 & 50.4 & 47.6 & 52   & 37.6 & 40.2 & 49.7 \\
RS-Net\cite{hassan2023regular} & 44.7 & 48.4 & 44.8 & 49.7 & 49.6 & 58.2 & 47.4 & 44.8 & 55.2 & 59.7 & 49.3 & 46.4 & 51.4 & 38.6 & 40.6 & 48.6 \\
Di2Pose\cite{wangdi} & \textcolor{red}{41.9}& 47.8 & 45.0 & 49.0 & 51.5 & 62.2 & 45.7 & 45.6 & 57.6 & 67.1 & 50.1 & 45.3 & 51.4 & \textcolor{blue}{37.3} & 40.9 & 49.2 \\
LiftingByImage\cite{zhou2024lifting}\textdaggerdbl & 44.9 & 46.4 & 42.4 & \textcolor{red}{44.9}  & 48.7  & \textcolor{red}{40.1}  & 44.3 & 55   & 58.9 & \textcolor{red}{47.1} & 48.2  & \textcolor{red}{42.6 }& \textcolor{red}{36.9}  & 48.8 & \textcolor{blue}{40.1}  & \textcolor{blue}{46.4} \\
GraphMLP\cite{li2025graphmlp} & 43.7 & 49.3 & 45.5 & 47.9 & 50.5 & 56.0 & 46.3 & \textcolor{red}{44.1} & 55.9 & 59.0 & 48.4 & 45.7 & \textcolor{blue}{51.2} & \textcolor{red}{37.1} & \textcolor{red}{39.1} & 48.0 \\
\hline
Ours ($H=1,K=1$)  & 44.1 & \textcolor{blue}{44.8} & \textcolor{blue}{41.5} & 48.6 & \textcolor{blue}{45.4} & 53.2 & \textcolor{blue}{44.2} & 45.2 & \textcolor{blue}{48.2} & 58.5 & \textcolor{blue}{46.0} & 45.5 & 53.1 & 41.0 & 43.1 & 46.8 \\
Ours ($H=10,K=5$) & 42.8 & \textcolor{red}{44.0} & \textcolor{red}{40.8} & \textcolor{blue}{47.8} & \textcolor{red}{44.3} & \textcolor{blue}{52.0} & \textcolor{red}{43.4} & \textcolor{blue}{44.4} & \textcolor{red}{47.4} & \textcolor{blue}{57.9} & \textcolor{red}{45.2} & \textcolor{blue}{44.8} & 52.1 & 40.4 & 41.8 & \textcolor{red}{46.0} \\
\hline
\hline
GraphSH\cite{xu2021graph} & 35.8 & 38.1 & 31.0 & 35.3 & 35.8 & 43.2 & 37.3 & 31.7 & 38.4 & 45.5 & 35.4 & 36.7 & 36.8 & 27.9 & 30.7 & 35.8 \\
HGN\cite{li2021hierarchical} & 35.4 & 40.2 & 31.1 & 38.2 & 38.3 & 41.1 & 36.1 & 32.7 & 42.1 & 48.4 & 37.1 & 36.9 & 37.1 & 30.5 & 32.4 & 37.2 \\
PHGANet\cite{zhang2023learning} & 32.4 & 36.5 & 30.1 & 33.3 & 36.3 & 43.5 & 36.1 & 30.5 & 37.5 & 45.3 & 33.8 & 35.1 & 35.3 & 27.5 & 30.2 & 34.9 \\
DiffPose\cite{gong2023diffpose}  & \textcolor{blue}{28.8} & 32.7 & 27.8 & 30.9 & 32.8 & 38.9 & \textcolor{blue}{32.2} & \textcolor{blue}{28.3} & 33.3 & 41.0 & 31.0 & 32.1 & 31.5 & 25.9 & 27.5 & 31.6 \\
LiftingByImage\cite{zhou2024lifting}\textdaggerdbl & 29.5 & \textcolor{red}{30.1} & 25.0 & 29.0 & 28.5 & \textcolor{red}{28.6} & \textcolor{red}{26.9} & 30.5 & 31.1 & \textcolor{red}{27.7} & 32.4 & \textcolor{red}{27.7} & \textcolor{red}{24.8} & 30.0 & 25.9 & \textcolor{blue}{28.6} \\
GraphMLP\cite{li2025graphmlp} & 32.2 & 38.2 & 29.3 & 33.4 & 33.5 & 38.1 & 38.2 & 31.7 & 37.3 & 38.5 & 34.2 & 36.1 & 35.5 & 28.0 & 29.3 & 34.2 \\
\hline
PoseFormer\cite{zheng20213d}($T=81$) & 30.0 & 33.6 & 29.9 & 31.0 & 30.2 & 33.3 & 34.8 & 31.4 & 37.8 & 38.6 & 31.7 & 31.5 & \textcolor{blue}{29.0} & \textcolor{blue}{23.3} & \textcolor{blue}{23.1} & 31.3 \\
MHFormer\cite{li2022mhformer}($T=351$) & \textcolor{red}{27.7} & 32.1 & 29.1 & \textcolor{blue}{28.9} & 30.0 & 33.9 & 33.0 & 31.2 & 37.0 & 39.3 & 30.0 & 31.0 & 29.4 & \textcolor{red}{22.2} & \textcolor{red}{23.0} & 30.5 \\
\hline
Ours ($H=1,K=1$)  & 29.9 & 31.9 & \textcolor{blue}{24.0} & 29.1 & \textcolor{blue}{27.4} & 29.5 & 33.5 & 28.4 & \textcolor{blue}{27.4} & 29.1 & \textcolor{blue}{27.5} & 30.1 & 29.6 & 24.8 & 26.6 & \textcolor{blue}{28.6} \\
Ours ($H=10,K=5$) & 29.7 & \textcolor{blue}{31.3} & \textcolor{red}{23.5} & \textcolor{red}{28.6} & \textcolor{red}{27.1} & \textcolor{blue}{28.9} & 33.0 & \textcolor{red}{27.8} & \textcolor{red}{27.1} & \textcolor{blue}{28.3} & \textcolor{red}{27.2} & \textcolor{blue}{29.4} & 29.2 & 24.6 & 26.3 & \textcolor{red}{28.1}\\
\hline
\end{tabular}
}
\label{table:h36m_cpn}
\end{table*}

\subsection{Training Objective}
To ensure the effective learning of spatial information, we supervise the framework with the Mean Squared Error (MSE) loss between the ground truth and estimated pose, which can be formulated as:
\begin{equation}
    \mathcal{L} = \lVert y_0 - \hat{y}_0 \rVert_2
\end{equation}
where $y_0$ and $\hat{y}_0$ are the ground truth and estimated 3D pose, respectively.
\section{Experiments}
\subsection{Datasets and Metrics}
We evaluate our method on two widely used datasets, \textbf{Human3.6M (H36M)} \cite{ionescu2013human3} and \textbf{MPI-INF-3DHP (3DHP)}. H36M is one of the most commonly used large-scale indoor 3D HPE datasets. It has 3.6 million video frames and encompasses a variety of 15 distinct activities performed by 11 professional subjects, captured at a 50 Hz frame rate through four synchronized and calibrated cameras. Following previous works \cite{shan2023diffusion,xu2024finepose}, our model is trained on five subjects (S1, S5, S6, S7, S8) and is evaluated on two subjects (S9, S11). We report the Mean Per Joint Position Error (MPJPE) and Procrustes-aligned Mean Per Joint Position Error (P-MPJPE) on this benchmark. 3DHP is also a public large-scale dataset. This dataset consists of 8 actors across 8 activities in the training subset and 7 activities in the evaluation subset under three distinct settings: green screen, non-green screen, and outdoor environments. Following \cite{zheng20213d,tang20233d,peng2024ktpformer}, we calculate MPJPE, the Percentage of Correct Keypoints (PCK) at a 150 mm threshold, and the Area Under the Curve (AUC) .

\subsection{Quantitative Results}

\begin{table}[t]
\caption{Performance comparison with state-of-the-art single-frame methods on MPI-INF-3DHP.}
\centering
\begin{tabular}{l|ccc}
\hline
Method                                & PCK↑ & AUC↑ & MPJPE↓ \\ \hline
Simple\cite{martinez2017simple} & 82.6 & 50.2 & 88.6   \\
Cascaded\cite{li2020cascaded} & 81.2 & 46.1 & 99.7   \\
MGCN\cite{zou2021modulated}     & 86.1 & 53.7 & -   \\
POT\cite{li2023pose} & 84.1 & 53.7 & -   \\
LiftingByImage\cite{zhou2024lifting}\textdaggerdbl   & \textcolor{blue}{88.2} & \textcolor{red}{59.3} & \textcolor{blue}{68.9}   \\ 
GraphMLP & 87.0 & 54.3 & - \\ 
\hline
Ours (\(H=1, K=1\)) & 87.6 & 57.0 & 69.2       \\ 
Ours (\(H=20, K=10\)) & \textcolor{red}{88.4} & \textcolor{blue}{58.7} & \textcolor{red}{68.5}       \\ 
\hline
\end{tabular}  
\label{table:mpi}
\end{table}

\begin{table}[t]
\caption{Comparison of computational complexity and efficiency.}
\centering
\scalebox{0.90}{
\begin{tabular}{l|cccccc}
\hline
Method    & H & K & MPJPE & Param (M) & FLOPs (G) & FPS\\
\hline
Graformer\cite{zhao2022graformer} & 1  & 1  & 51.8      & 0.65      & 0.702         & 21588 \\
GraphMLP\cite{li2025graphmlp}     & 1  & 1  & 48.0      & 9.49      & 0.348    & 41143  \\
D3DP\cite{shan2023diffusion}      & 1  & 1  & 48.9      &  34.71    & 1.152    & 15514   \\
\hline
Ours      &1   & 1 & 46.8 &  13.07     & 0.443          &  30289        \\
Ours      & 5  & 5 & \textcolor{blue}{46.1} &  13.07     & 11.053         &   866       \\
Ours      & 10 & 5 & \textcolor{red}{46.0} &  13.07     & 22.102         &   412      \\
\hline
\end{tabular}
}
\label{table:efficiency}
\end{table}

\subsubsection{\textbf{H36M}} 
As shown in Table~\ref{table:h36m_cpn}, we compare the proposed HyperDiff with state-of-the-art  methods on the Human3.6M dataset. When using detected 2D poses and ground truth 2D poses as inputs, our method achieves $46.8$ mm and $28.6$ mm ($H=1, K=1$), respectively, ranking second only to the LiftingByImage~\cite{zhou2024lifting}, which incorporates additional visual information. Furthermore, when applying the multi-hypothesis strategy from D3DP\cite{shan2023diffusion}, our model outperforms LiftingByImage in terms of accuracy ($H=10,K=5$). These results demonstrate that combining diffusion models with HyperGCN effectively captures high-order dependencies between joints, thereby significantly enhancing pose estimation accuracy.

\subsubsection{\textbf{3DHP}}
As shown in Table~\ref{table:mpi}, our proposed method outperforms existing state-of-the-art approaches on the MPI-INF-3DHP dataset, particularly in the PCK and AUC metrics. With the $H=20, K=10$ configuration, our model achieves $88.4\%$ PCK and $58.7\%$ AUC, surpassing LiftingByImage and other comparative methods. These results demonstrate higher accuracy and greater generalization ability, indicating stable performance in diverse scenarios.
\subsubsection{\textbf{Efficiency}}
As shown in Table~\ref{table:efficiency}, our method demonstrates superior computational complexity and efficiency. With $H=1, K=1$, and fewer model parameters and FLOPs than D3DP, our model achieves a faster inference speed of $30289$ FPS while maintaining the MPJPE of $46.8$ mm. Although there is a decrease in speed at larger configurations, our method still outperforms others in terms of precision. Thus, it effectively balances accuracy and efficiency, making it suitable to be adapted to real-time applications.

\subsection{Qualitative Results}

\begin{figure}
	\centering
    \includegraphics[width=\linewidth]{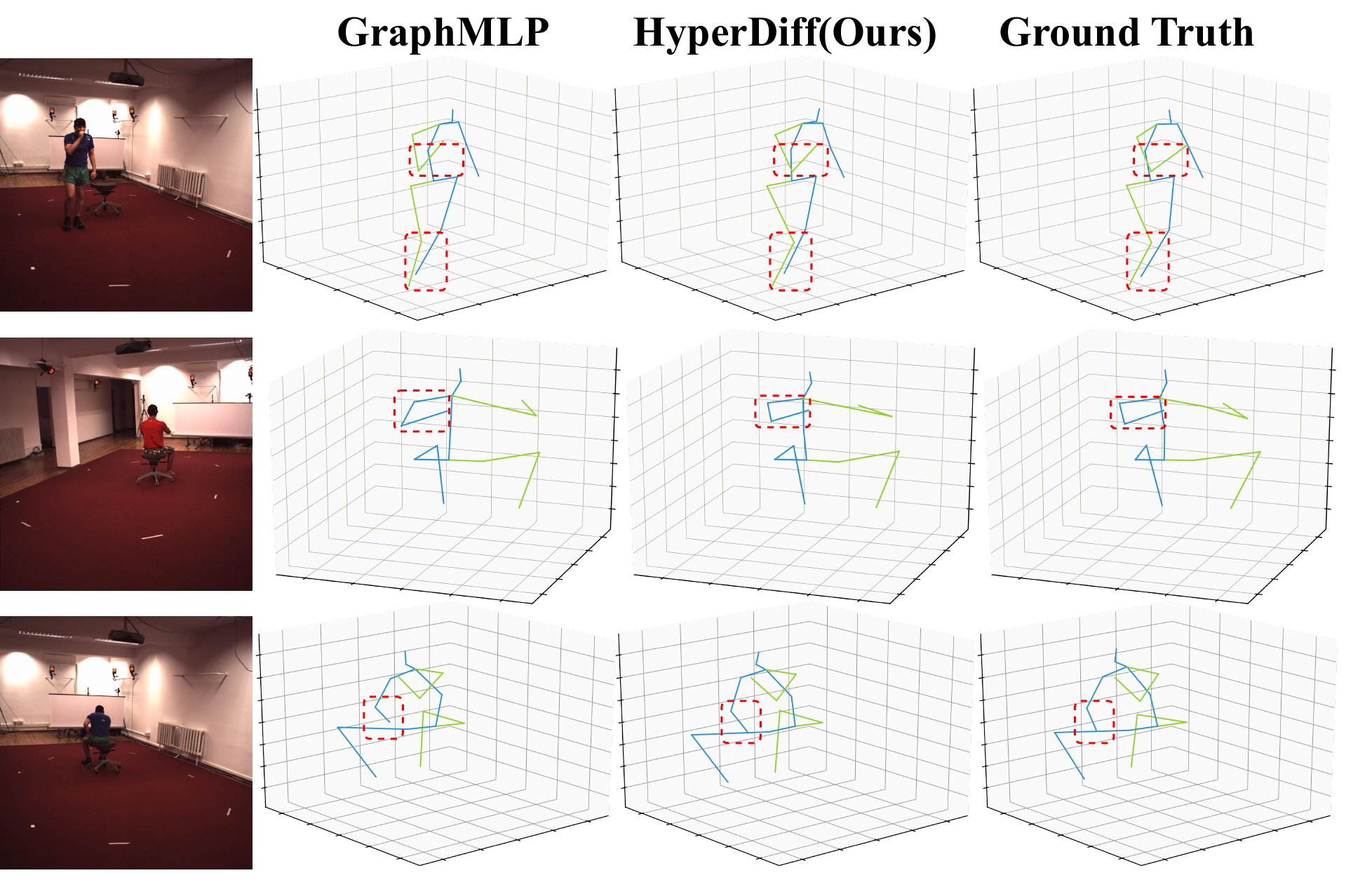} \\ 
    \vspace{-2mm}
	\caption{Qualitative results on Human3.6M. The blue/green line represents the estimated pose sequence for the left/right skeleton.}
	\label{fig:comparison} 
    \vspace{-4mm}
\end{figure}
\begin{figure}[t]
	\centering
    \includegraphics[width=\linewidth]{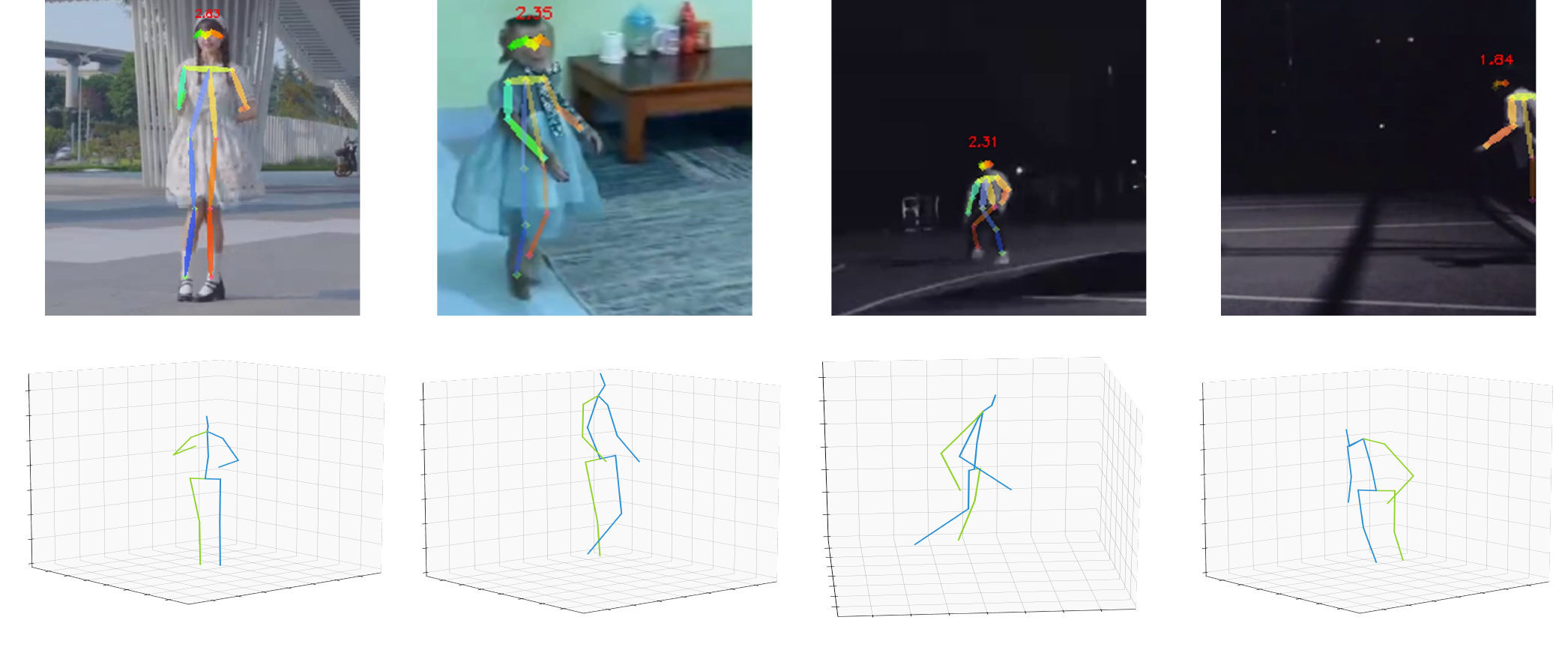} \\ 
    \vspace{-2mm}
	\caption{Qualitative results of our method on in-the-wild videos with 3D human poses under diverse and challenging scenarios, including depth ambiguity, self-occlusion, faulty 2D detection, and incomplete 2D pose.}
	\label{fig:wild} 
    \vspace{-4mm}
\end{figure}

Fig.~\ref{fig:comparison} presents the qualitative comparison of HyperDiff with GraphMLP\cite{li2025graphmlp} on the Human3.6M dataset. Leveraging the multi-scale skeleton structure, our approach more effectively handles complex scenarios, such as self-occlusion. Additionally, Fig.~\ref{fig:wild} presents the qualitative results of HyperDiff on more challenging in-the-wild images to assess the generalization ability of our model. It is important to note that the actions in these natural videos are relatively rare or absent in the training set. For example, the first instance exhibits severe depth ambiguity in the hand, the second involves self-occlusion, and the last two show detection errors and incomplete detections from the 2D detector. Despite these challenges, our method demonstrates strong generalization, accurately predicting 3D poses by leveraging the learned skeleton feature.

\begin{table}
    \centering
    \begin{minipage}[t]{0.5\linewidth}
        \caption{Ablation study on different-granularity graph combinations.}
        \centering
        \begin{tabular}{l|c}
            \hline
            \textbf{Configuration} & MPJPE$\downarrow$  \\ 
            \hline
            Baseline  & 48.9     \\ 
            \hline
            Joint-scale  &  49.5\textcolor{blue}{(+0.6)}   \\ 
            +Part-scale &  48.2\textcolor{red}{(-0.7)}  \\ 
            +Body-scale &  47.6\textcolor{red}{(-1.3)} \\ 
            +Part-scale+Body-scale & \textbf{46.8}\textcolor{red}{(-2.1)}    \\ 
            \hline
        \end{tabular}
        \label{tab:graph_scale}
    \end{minipage} \hfill
    \begin{minipage}[t]{0.45\linewidth}
        \caption{Ablation study on feature fusion strategy.}
        \centering
        \begin{tabular}{l|c}
            \hline
            \textbf{Fusion Strategy} & MPJPE$\downarrow$ \\
            \hline
            Concate Fusion & 49.4 \\
            Product Fusion & 47.1 \\
            Weighted Fusion & \textbf{46.8} \\
            \hline
        \end{tabular}
        \label{tab:fusion_strategy}
    \end{minipage}
\end{table}

\subsection{Ablation Study}
\subsubsection{\textbf{Different Combinations of Graph-scale}}
As shown in Table~\ref{tab:graph_scale}, incorporating graph structures at different levels significantly improves the performance of the model. Compared to the baseline (D3DP at $T=1$), using only the joint-scale graph increases the MPJPE to $49.5$ mm, suggesting that joint-level graphs alone have a limited impact on performance. When the part-scale and body-scale graphs are added, the MPJPE decreases by $0.7$ mm and $1.3$ mm, respectively, indicating that higher-level structures better model inter-joint relationships. Notably, when both the part-scale and body-scale graphs are used together, the MPJPE reaches its lowest value of $46.8$ mm. That demonstrates that combining local and global structures allows for a more comprehensive capture of complex joint dependencies, significantly enhancing pose estimation accuracy.



\subsubsection{\textbf{Feature Fusion Strategy}}
Table~\ref{tab:fusion_strategy} illustrates the impact of different feature fusion strategies on model performance. Concatenate fusion yields the poorest results, as basic concatenation fails to effectively integrate multi-level graph features. In contrast, weighted fusion achieves the best performance by assigning weights to different features according to their importance, enabling the model to better exploit the advantages of multi-level graphs. While product fusion also captures feature interactions effectively, it is slightly less precise than weighted fusion in fusing multi-level features.

\section{Conclusion}
This study presents HyperDiff, a method that leverages diffusion models with HyperGCN to address depth ambiguity and self-occlusion in 3D human pose estimation. By incorporating the multi-granularity hypergraph GCN, HyperDiff enhances the modeling of high-order dependencies between joints, thereby improving pose estimation accuracy. Experimental results demonstrate that HyperDiff achieves exceptional performance on multiple standard datasets. Additionally, it effectively balances computational efficiency with performance, making it suitable for real-time applications.

\bibliographystyle{ieeetr}
\bibliography{ref}

\end{document}